 \documentclass[pmlr,twocolumn,10pt]{jmlr} % W&CP article

\usepackage{booktabs}
\usepackage{siunitx}

\usepackage[switch]{lineno}

\theorembodyfont{\upshape}
\theoremheaderfont{\scshape}
\theorempostheader{:}
\theoremsep{\newline}

\jmlrworkshop{Machine Learning for Health (ML4H) 2025} % W&CP title

\title[Cancer-Net PCa-MultiSeg: CDI$^s$-Enhanced Prostate Cancer Lesion Segmentation]{Cancer-Net PCa-MultiSeg: Multimodal Enhancement of Prostate Cancer Lesion Segmentation Using Synthetic Correlated Diffusion Imaging}

 \author{%
 \Name{Jarett Dewbury} \Email{jdewbury@uwaterloo.ca}\\
 \Name{Chi-en Amy Tai} \Email{amy.tai@uwaterloo.ca}\\
 \Name{Alexander Wong} \Email{alexander.wong@uwaterloo.ca}\\
 \addr Vision and Image Processing Group, Systems Design Engineering,
University of Waterloo
 }

\begin{document}

\maketitle

\begin{abstract}
Current deep learning approaches for prostate cancer lesion segmentation achieve limited performance, with Dice scores of 0.32 or lower in large patient cohorts. To address this limitation, we investigate synthetic correlated diffusion imaging (CDI$^s$) as an enhancement to standard diffusion-based protocols. We conduct a comprehensive evaluation across six state-of-the-art segmentation architectures using 200 patients with co-registered CDI$^s$, diffusion-weighted imaging (DWI) and apparent diffusion coefficient (ADC) sequences. We demonstrate that CDI$^s$ integration reliably enhances or preserves segmentation performance in 94\% of evaluated configurations, with individual architectures achieving up to 72.5\% statistically significant relative improvement over baseline modalities. CDI$^s$ + DWI emerges as the safest enhancement pathway, achieving significant improvements in half of evaluated architectures with zero instances of degradation. Since CDI$^s$ derives from existing DWI acquisitions without requiring additional scan time or architectural modifications, it enables immediate deployment in clinical workflows. Our results establish validated integration pathways for CDI$^s$ as a practical drop-in enhancement for PCa lesion segmentation tasks across diverse deep learning architectures.

\end{abstract}

\paragraph*{Data and Code Availability}
This study leveraged the Cancer-Net PCa-Data data set~\citep{gunraj2023cancer}, a publicly available data set designed for multimodal PCa analysis. The data set was derived from the SPIE-AAPM-NCI PROSTATEx Challenges, PROSTATEx\_masks, and The Cancer Imaging Archive (TCIA) datasets~\citep{Litjens2014,Litjens2017,Cuocolo2021,Clark2013}. Code is available at \url{https://github.com/Jdewbury/Cancer-Net-PCa-MultiSeg}

\paragraph*{Institutional Review Board (IRB)}
This research does not require IRB approval, as it exclusively uses publicly available data that have previously been approved for research use.
\vspace{-10px}

\begin{figure*}[t]
  \centering
  \includegraphics[width=0.85\linewidth]{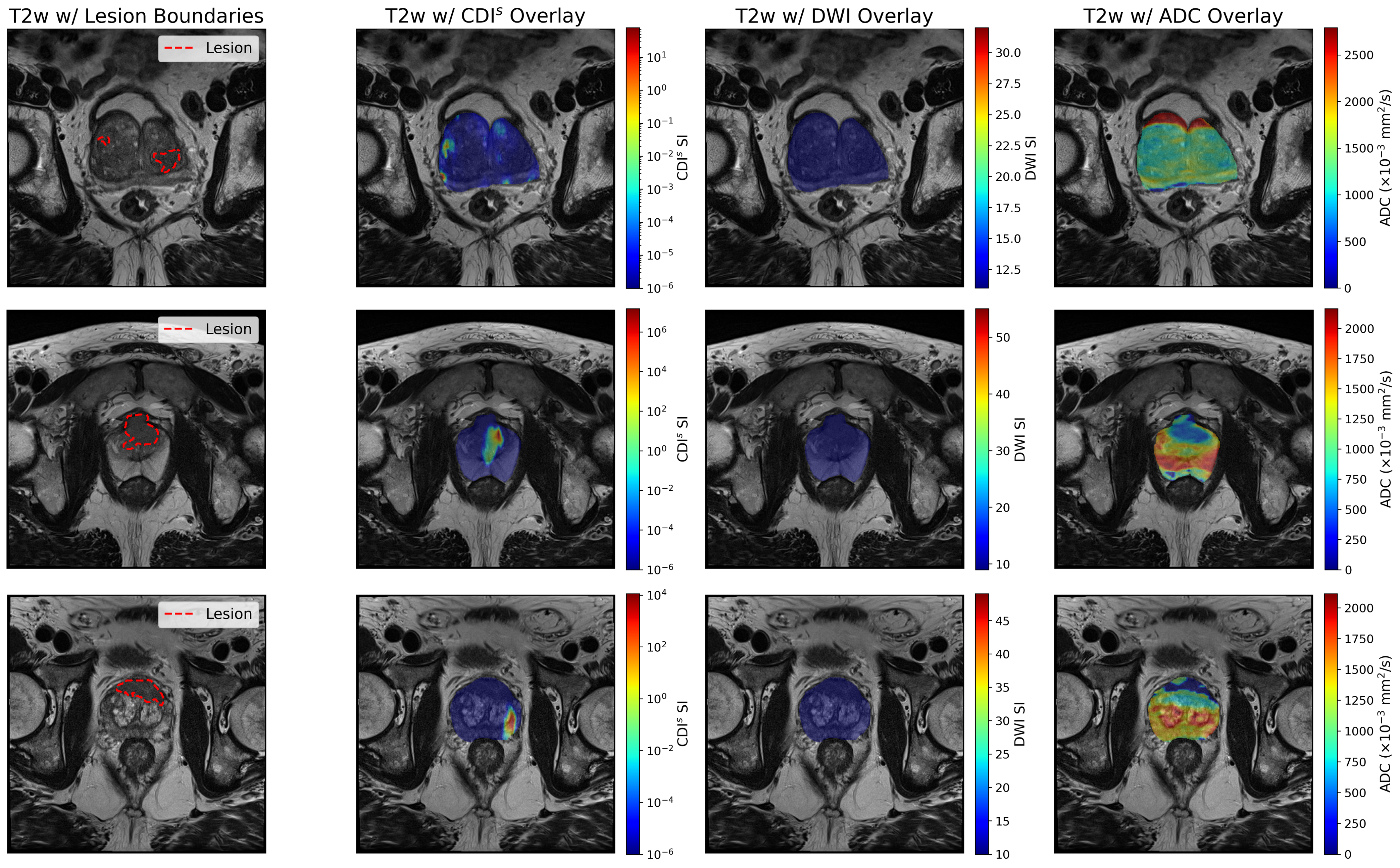}
  \caption{Visualization of sample T2-weighted (T2w) images with overlays highlighting lesion boundaries (red dashed lines) and comparative intensity distributions for CDI$^s$, DWI, and ADC sequences. We consider both clinically significant (csPCa) and clinically insignificant (insPCa) lesions.}
  \label{fig:figure_1}
  \vspace{-10px}
\end{figure*}

\section{Introduction}
\label{sec:intro}

Prostate cancer (PCa) is the leading cancer diagnosis among men in the United States, accounting for more than 310,000 new cases and 35,000 total deaths in 2025 ~\citep{siegel2025cancer}. Accurate localization and detection of PCa lesions remain fundamental in guiding treatment protocols and improving patient outcomes. Multiparametric magnetic resonance imaging (mpMRI) has established itself as the primary diagnostic modality for the evaluation of PCa, allowing non-invasive visualization of prostatic tissue and serving as the basis for standardized assessment protocols such as the Prostate Imaging Reporting and Data System (PI-RADS) ~\citep{turkbey2019prostate}. 

Despite technological advances, current automated approaches for PCa lesion detection exhibit significant performance constraints. Deep learning methods applied to mpMRI data for lesion segmentation demonstrate limited efficacy, with recent investigations reporting Dice score values of 0.32 ~\citep{gunashekar2022explainable} and 0.276 ~\citep{pellicer2022deep} on large patient cohorts. These results highlight the persistent challenges of accurately differentiating malignant and benign prostatic tissue, emphasizing the need for improved imaging techniques that offer enhanced diagnostic precision. However, such approaches must also maintain clinical feasibility by balancing gains in diagnostic accuracy with reasonable computational demands and minimal modifications to established workflows. 

Synthetic correlated diffusion imaging (CDI$^s$) presents a promising approach to address current limitations in PCa lesion detection. CDI$^s$ leverages existing diffusion-weighted acquisitions to enhance tissue contrast while maintaining compatibility with established mpMRI protocols and requiring no additional scan time. This technique demonstrates superior delineation between cancerous and healthy prostatic tissue, as well as fewer false positives compared to standard mpMRI techniques ~\citep{wong2022synthetic}. 

In this study, we investigate the integration of CDI$^s$ with standard diffusion-weighted imaging (DWI) and apparent diffusion coefficient (ADC), which constitute the diffusion component of contemporary mpMRI protocols. DWI serves as the dominant sequence for peripheral zone assessment in the PI-RADS evaluation, while ADC provides complementary diffusion characterization ~\citep{turkbey2019prostate}. The systematic evaluation of CDI$^s$ enhancement in various deep learning architectures for PCa lesion segmentation remains unexplored. This work presents the first comprehensive benchmark evaluating CDI$^s$ integration with standard DWI and ADC sequences in six state-of-the-art medical segmentation models, demonstrating improved lesion detection performance and establishing a framework for integrating advanced imaging techniques into existing clinical workflows to enhance PCa diagnostic capabilities.

\vspace{-10px}
\section{Methodology}
\label{sec:method}

\begin{figure*}[t]
  \centering
  \includegraphics[width=\linewidth]{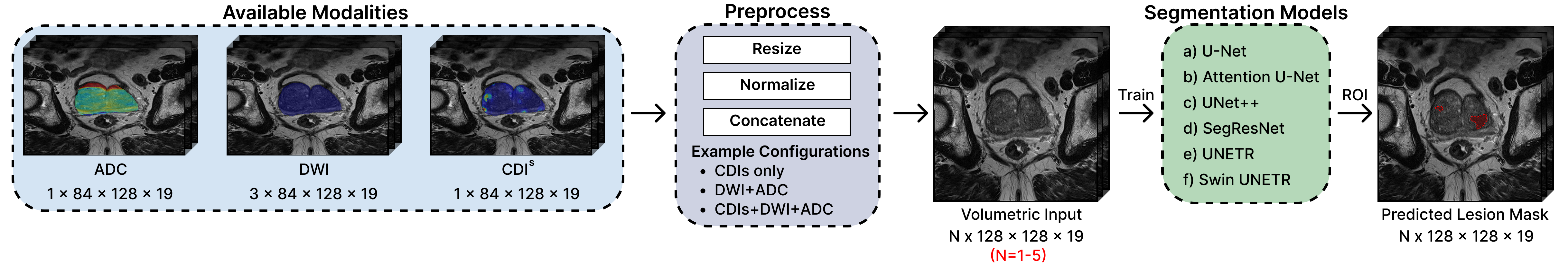}
  \caption{Multimodal prostate cancer lesion segmentation pipeline. The workflow includes modular input volume selection from three modalities (CDI$^s$, DWI, ADC), standardized preprocessing, evaluation across six segmentation architectures, and binary lesion mask prediction output.}
  \label{fig:figure_2}
    \vspace{-10px}
\end{figure*}

\subsection{Dataset}

 This study uses the Cancer-Net PCa-Data data set comprising volumetric imaging data from a cohort of 200 patients, including three spatially co-registered diffusion-based sequences: CDI$^s$, DWI and ADC (see Figure~\ref{fig:figure_1} for sample patient data). These modalities share identical spatial coordinates and voxel dimensions, allowing for direct pixel-wise comparison and eliminating registration-related artifacts that could confound multimodal analysis.
 Ground truth annotations include lesion masks validated by expert radiologists, encompassing both clinically significant and insignificant PCa lesions defined by Gleason scoring criteria. 

 \textbf{CDI$^s$ Computation.} CDI$^s$ is derived from standard DWI acquisitions without requiring additional scan time ~\citep{wong2022synthetic}. The computation involves three steps: (1) multiple DWI signals are acquired at different b-values; (2) additional signals at higher b-values are synthesized from native acquisitions using a signal synthesizer, avoiding lengthy high b-value acquisitions that would increase scan time and introduce  artifacts; (3) native and synthetic signals are combined through a correlation-based mixing function that analyzes signal intensity co-variation across different b-values. This process preserves information on the distribution of water mobility within cancerous regions that exhibit high signal across all b-values, while healthy tissue shows high signal only at low b-values. Unlike ADC, which reduces diffusion to a single number, CDI$^s$ retains distributional information to enable better differentiation of malignant tissue. Since CDI$^s$ utilizes existing DWI sequences, it integrates seamlessly into clinical workflows without protocol modifications.

\vspace{-10px}
\subsection{Experimental Setup}

Our experimental pipeline (Figure~\ref{fig:figure_2}) facilitates a systematic evaluation of the enhancement of CDI$^s$ in multiple architectures and combinations of modalities. Patient-level stratification was applied to prevent data leakage, with 90\% allocated to training/validation and 10\% to the held-out test set. The training/validation subset was further partitioned using 10-fold cross-validation, enabling robust performance estimation across diverse patient populations. All modality and ground-truth lesion mask volume pairs were standardized to 128x128x19 voxels using bilinear interpolation for images and nearest-neighbor interpolation for masks. Channel-wise min-max normalization was applied to the modality volumes to ensure uniform intensity scaling in the [0, 1] range. 

Seven modality configurations were systematically evaluated to assess CDI$^s$ enhancement across three categories: individual modality (CDI$^s$, ADC, DWI), standard clinical protocol (DWI + ADC), and CDI$^s$-enhanced combinations (CDI$^s$ + ADC, CDI$^s$ + DWI, CDI$^s$ + DWI + ADC). This design enables direct assessment of CDI$^s$ performance contributions when integrated with standard diffusion sequences while establishing individual modality baselines for PCa lesion segmentation.

\begin{table*}[!t]
\floatconts
  {tab:main_results}%
  {\caption{Quantitative comparisons of prostate lesion segmentation across imaging modalities and architectures. Dice score (mean $\pm$ standard deviation) on the held-out test set. Best performance per architecture is indicated in bold. Statistical significance from paired t-tests: $^{*}$p$<$0.05, $^{**}$p$<$0.01.}\vspace{-10px}}%
  {\footnotesize\setlength{\tabcolsep}{4pt}\begin{tabular}{lcccccc}
  \toprule
  \textbf{Modality} & \textbf{U-Net} & \textbf{Attention U-Net} & \textbf{UNet++} & \textbf{SegResNet} & \textbf{UNETR} & \textbf{Swin UNETR} \\
  \midrule
  CDI$^s$ & $24.64 \pm 3.1$ & $26.72 \pm 1.4$ & $27.27 \pm 1.5$ & $27.58 \pm 1.4$ & $26.34 \pm 1.3$ & $26.86 \pm 3.1$ \\
  ADC & $17.47 \pm 3.2$ & $25.50 \pm 4.0$ & $31.58 \pm 1.7$ & $30.74 \pm 3.1$ & $35.04 \pm 1.7$ & $31.78 \pm 2.5$ \\
  DWI & $21.55 \pm 3.3$ & $14.82 \pm 2.8$ & $29.27 \pm 2.4$ & $29.39 \pm 2.1$ & $31.49 \pm 2.1$ & $30.48 \pm 1.9$ \\
  \midrule
  DWI+ADC & $22.31 \pm 3.1$ & $20.70 \pm 4.0$ & $31.05 \pm 2.9$ & $\mathbf{30.96 \pm 2.1}$ & $\mathbf{35.08 \pm 2.3}$ & $33.15 \pm 2.1$ \\
  \midrule
  CDI$^s$+ADC & $\phantom{^{**}}23.14 \pm 2.1^{**}$& $\mathbf{28.55 \pm 2.0}$ & $31.70 \pm 1.6$ & $\phantom{^{*}}28.53 \pm 2.2^{*}$& $34.16 \pm 1.0$ & $32.82 \pm 1.8$ \\
  CDI$^s$+DWI & $\mathbf{\phantom{^{**}}26.68 \pm 1.5}^{**}$& $\phantom{^{**}}25.57 \pm 5.1^{**}$& $29.75 \pm 1.6$ & $29.44 \pm 2.4$ & $32.34 \pm 1.0$ & $\phantom{^{*}}32.56 \pm 2.2^{*}$\\
  CDI$^s$+DWI+ADC & $\phantom{^{*}}25.70 \pm 2.5^{*}$& $24.44 \pm 4.0$ & $\mathbf{32.70 \pm 2.3}$ & $30.51 \pm 2.1$ & $34.54 \pm 2.0$ & $\mathbf{34.30 \pm 1.8}$ \\
  \bottomrule
  \end{tabular}}
\end{table*}

\vspace{-10px}
\subsection{Model Architectures}

Six state-of-the-art medical segmentation architectures were evaluated: SegResNet~\citep{myronenko20183dmribraintumor}, U-Net~\citep{ronneberger2015unetconvolutionalnetworksbiomedical}, UNet++~\citep{zhou2018unet++}, Attention U-Net~\citep{oktay2018attentionunetlearninglook}, UNETR~\citep{hatamizadeh2022unetr}, Swin UNETR~\citep{hatamizadeh2022swinunetrswintransformers}. All models were implemented using the MONAI framework ~\citep{cardoso2022monai} with standardized hyperparameters and preprocessing pipelines to ensure fair comparison across architectures. This controls implementation-specific optimizations that could confound modality contribution analysis. This experimental design yields 42 architecture-modality configurations (6 architectures x 7 modality combinations), of which 18 constitute direct CDI$^s$ enhancement comparisons against baseline modalities, enabling comprehensive evaluation of CDI$^s$ benefits across diverse network designs and input combinations. 

\vspace{-10px}
\subsection{Training Protocol}

Models were trained using binary cross-entropy with logits loss and AdamW optimizer with ReduceLROnPlateau scheduling. Training employed early stopping based on validation Dice score with a patience of 50 epochs and minimum improvement threshold of 0.001. Model outputs underwent sigmoid activation followed by optimal probability thresholding to generate binary segmentation masks. Optimal thresholds were determined per fold using systematic grid search over [0.1, 1.0] with 0.01 increments, selecting the threshold value that maximized validation Dice performance for test set evaluation.

\vspace{-10px}
\section{Results}
\label{sec:results}

\tableref{tab:main_results} presents a comparative segmentation performance analysis across all evaluated modality configurations and architectures. Results are reported on the held-out testing set as the mean Dice score ± standard deviation across 10-fold cross-validation.

\textbf{Statistical Significance Testing.} To assess the reliability of observed performance differences, we conducted paired t-tests across 10-fold cross-validation comparing each CDI$^s$-enhanced configuration against its corresponding baseline. Of 18 direct CDI$^s$ enhancement comparisons, 14 showed improved mean Dice scores, among which 5 achieved statistical significance. One configuration showed significant degradation (SegResNet + CDI$^s$ + ADC), while 12 maintained baseline performances without significant differences. This demonstrates that CDI$^s$ reliably enhances or preserves segmentation performance in 94\% of the evaluated configurations. 

\textbf{Configuration-Level Analysis.} CDI$^s$ + DWI emerged as the most reliable enhancement configuration, achieving significant improvements in 3 of 6 architectures (U-Net, Attention U-Net, Swin UNETR) with no degradations. Attention U-Net showed the most substantial gains, achieving a relative improvement of 72.5\% when adding CDI$^s$ to DWI (CDI$^s$ + DWI: $25.57 \pm 5.1$ vs. DWI alone: $14.82 \pm 2.8$). This configuration showed mean performance improvements across all six architectures, establishing it as a safe drop-in enhancement for existing DWI protocols. 

\textbf{Architecture-Level Analysis.} Convolutional neural network (CNN) architectures benefited the most from CDI$^s$ integration. U-Net demonstrated significant improvements across all three CDI$^s$ configurations, establishing it as the most reliable architecture for CDI$^s$ deployment with a maximum gain of +5.13. Attention U-Net achieved the largest performance gain (+10.75) in the CDI$^s$ + DWI configuration compared to baseline. Transformer architectures exhibited more selective enhancement patterns. Swin UNETR achieved significant improvement with CDI$^s$ + DWI and demonstrated better mean Dice gains across all three CDI$^s$ configurations compared to their respective baselines. UNETR showed no significant differences, despite maintaining high absolute performance across modality configurations. This suggests that global attention mechanisms effectively extract relevant features from standard diffusion modalities, resulting in diminishing returns from CDI$^s$ enhancement in their current state. Therefore, specialized architectural modifications or training strategies may be necessary to fully leverage CDI$^s$ information in these models. SegResNet demonstrated mixed compatibility, showing mean performance improvements with CDI$^s$ + DWI, but exhibiting the only significant degradation observed across all experiments when pairing CDI$^s$ with ADC alone. 

\vspace{-10px}
\section{Conclusion}
\label{sec:conclusion}
In this paper, we investigate the integration of CDI$^s$ with diffusion-based modalities utilized in mpMRI protocols. We conduct a comprehensive evaluation across six state-of-the-art medical segmentation architectures to assess CDI$^s$ enhancement capabilities for PCa lesion detection. We demonstrate that CDI$^s$ reliably enhances or preserves segmentation performance in 94\% of evaluated configurations without architectural modifications or specialized training procedures.

Our findings establish two critical insights for clinical deployment. First, CDI$^s$ + DWI represents the safest enhancement pathway, achieving significant improvements in half of the evaluated architectures with zero instances of performance degradation. Second, U-Net exhibits reliable CDI$^s$ compatibility across all modality combinations, establishing it as the recommended architecture for CDI$^s$-enhanced segmentation workflows. These validated enhancement pathways provide clear deployment guidance for clinical implementation. 

Since CDI$^s$ derives from standard DWI acquisitions already performed in clinical protocols, it requires no additional scan time or protocol modifications for integration. This makes CDI$^s$ enhancement immediately deployable in existing clinical workflows without infrastructure changes or additional patient burden. We acknowledge that absolute Dice scores remain modest and consistent with reported literature for PCa lesion segmentation, reflecting the inherent difficulty of this task across all evaluated architectures. Our primary contribution demonstrates statistically validated relative enhancements across diverse model architectures with clear, actionable deployment guidance. This zero-barrier integration distinguishes CDI$^s$ from alternative enhancement approaches that would require new hardware, protocol modifications, or extended scan times.

Future work should evaluate CDI$^s$ effectiveness across additional datasets, investigate architecture-specific optimization strategies to maximize transformer-based CDI$^s$ utilization, and assess clinical impact through prospective validation studies. CDI$^s$ represents a practical advancement in PCa lesion diagnostic workflows that can benefit detection systems with minimal integration barriers. 

\bibliography{jmlr-sample}
\end{document}